\documentclass[letterpaper, 10 pt, conference]{ieeeconf}  

\IEEEoverridecommandlockouts                              
\usepackage{amsmath} 
\usepackage[capitalise]{cleveref}
\usepackage{graphicx}
\usepackage{graphics} 
\usepackage{float} 
\usepackage{subfig}

\usepackage{cite}
\usepackage{amsmath,amssymb,amsfonts}
\usepackage{algorithmic}
\usepackage{textcomp}
\usepackage{xcolor}
\usepackage{algorithm}

\usepackage{eso-pic}

\newcommand{\IEEEcopyrightnotice}{%
  \AddToShipoutPictureFG*{%
    \AtPageLowerLeft{%
      \raisebox{4mm}{%
        \makebox[\paperwidth][c]{%
          \parbox{0.92\paperwidth}{%
            \centering
            \fontsize{6}{6.5}\selectfont
            \copyright~2025 IEEE. Personal use of this material is permitted. Permission from IEEE must be obtained for all other uses, in any current or future media, including reprinting/republishing this material for advertising or promotional purposes, creating new collective works, for resale or redistribution to servers or lists, or reuse of any copyrighted component of this work in other works.\\
            \textit{IEEE Robotics and Automation Letters}, vol.~10, no.~12, pp.~13003--13010, 2025. DOI: 10.1109/LRA.2025.3625507.
          }%
        }%
      }%
    }%
  }%
}

\title{\LARGE \bf
An End-to-End Framework for Modelling Pneumatic Soft Robots Based on Differentiable Finite Element Methods
}

\author{Shaohong Zhong$^{1}$, Yao Yao$^{1}$, Perla Maiolino$^{1}$ and Ingmar Posner$^{1}$% <-this % stops a space
\thanks{$^{1}$Oxford Robotics Institute, Department of Engineering Science, University of Oxford, OX1 3PJ Oxford, U.K}%}
}

\begin{document}

\IEEEcopyrightnotice

\maketitle
\thispagestyle{empty}
\pagestyle{empty}

\begin{abstract}

Soft robots present significant modelling challenges due to their non-linearity, complex dynamics and potentially intricate geometries. These difficulties in accurate system identification and dynamics modelling limit their applications in precise robotics tasks. Prior modelling approaches typically suffer from trade-offs in accuracy, computational efficiency, or speed. The differentiable finite element method (FEM) offers a promising balance between these desiderata for soft robot modelling, and enables the use of gradients for efficient calibration and trajectory optimisation. In this paper, we propose an end-to-end differentiable FEM-based framework designed to streamline modelling and system identification for pneumatic soft robots, enabling the generation of reliable models for downstream tasks. The framework automates the conversion of computer-aided designs into voxelised tetrahedral viscoelastic FEM meshes that accurately represent the robot's geometry and actuation mechanism. By integrating easily acquired point cloud data with differentiable FEM, the framework achieves precise identification of material and dynamic actuation parameters using a minimal experimental setup. Additionally, the differentiable nature of the model facilitates trajectory optimisation by leveraging gradients from robot dynamics and contact interactions. We validate the framework by modelling a complex bellow-shaped pneumatic soft robot and demonstrate its efficacy in real-world motion planning tasks. Experimental results indicate high modelling accuracy, with a maximum positional error of less than 3 mm, and successful application in tasks such as path following and grasping.

\end{abstract}

\section{INTRODUCTION}

Soft robots hold significant promise for achieving a high degree of adaptability and robustness, enabling safe manipulation of unknown objects and operation in unstructured environments~\cite{hughes2016soft,chen2022review}. By exploiting innovative designs and smart materials, soft robots can embody physical intelligence, allowing them to effectively respond to dynamic environmental challenges and uncertainties~\cite{mazzolai2022roadmap,kim2013soft}. Within this domain, pneumatic actuators play a key role and are widely used for their lightweight design, high power-to-weight ratio, and simple manufacturing~\cite{xaview2022soft}. 

For soft robots to reliably perform precise tasks, efficient and accurate models are required. However, modelling them is challenging due to material nonlinearities, complex dynamics, and intricate geometries~\cite{hughes2016soft, chen2022review, thuruthel2018stable}. These challenges are particularly pronounced in contact-rich scenarios, where robot dynamics couples with the environment~\cite{santina2023model}. These factors result in behaviours that are difficult to predict or computationally expensive to simulate without simplifying assumptions about geometry and deformation in prior analytical, numerical, or learning-based approaches~\cite{thuruthel2018stable,santina2023model,dubied2022sim,du2021underwater}. 

\begin{figure} [h!]
	\centering
    \includegraphics[width=0.6\columnwidth]{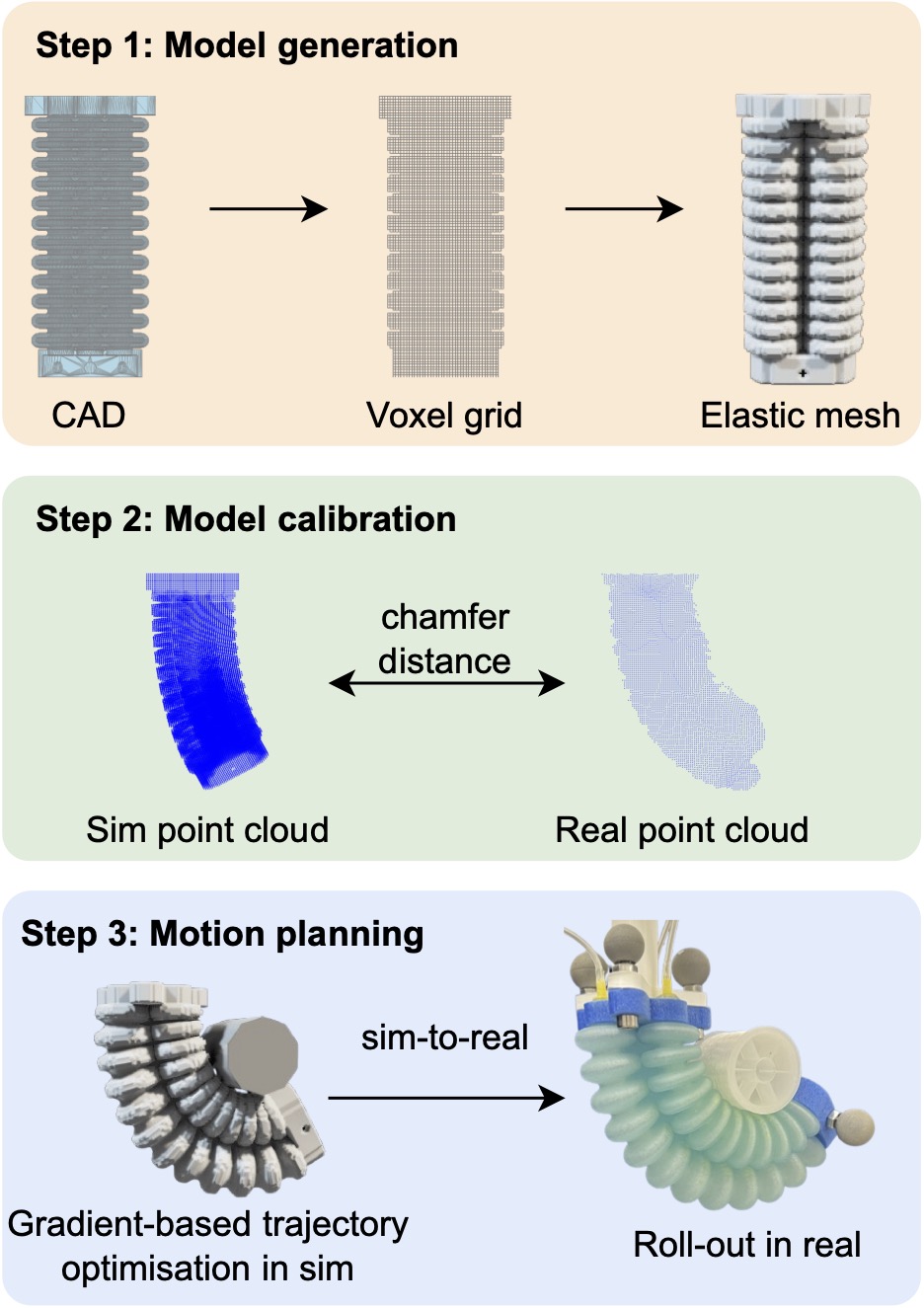}
	\caption{Overview of the framework. Given the CAD of a soft robot, we first generate a voxelised finite element model. Next, the material and dynamic actuation parameters are identified using point clouds. Following identification, the differentiable model enables gradient-based motion planning.}
	\label{fig:overview}
    \vspace{-1.5mm}
\end{figure} 

Compounding the challenges in dynamics modelling, the system identification process for soft robots, which is required for accurate modelling, also presents significant challenges. Past approaches are often laborious, requiring extensive human effort to design and tailor experimental setups for individual robots, and potentially require a substantial amount of data for calibration~\cite{hughes2016soft,chen2022review,shintake2018soft}. These challenges collectively hinder effective modelling of soft robots and limit their full potential in robotics tasks.
 
The differentiable finite element method (FEM) has emerged as an accurate and efficient approach for modelling soft robots~\cite{Bacher2021DesignControl}. By employing spatial discretisation, FEM numerically represents soft robots as a collection of interacting volumetric mesh elements, effectively capturing their nonlinear behaviours~\cite{dubied2022sim}. The method also enables direct gradient computation with respect to model parameters and control inputs, facilitating parameter estimation through real-world observations and other downstream tasks~\cite{dubied2022sim, du2021underwater, zhang2022sim}. 

In this work, we propose an end-to-end framework for efficient model creation and system identification for pneumatic soft robots using differentiable FEM, demonstrated on a complex bellow-shaped pneumatic actuator. An overview is in \cref{fig:overview}. To minimise manual efforts in model creation, we introduce a generation method that directly converts CAD models to volumetric meshes via voxelisation. For model calibration, inspired by~\cite{sundaresan2022diffcloud}, we leverage point cloud data from RGB-D cameras and perform gradient-based optimisation of material and dynamic actuation parameters using GPU-accelerated differentiable simulation. Furthermore, we demonstrate the utility of the model through a series of downstream real-world motion planning tasks.

\section{LITERATURE REVIEW}

This section starts with a brief overview of prior approaches in soft robot modelling to highlight the current challenges. It follows with a review of existing exploratory works using differentiable FEM. 

\subsection{Soft Robot Modelling}

Soft robot modelling approaches can be broadly classified into three categories. The first involves analytical models that rely on simplified assumptions about geometry and deformation, such as Piecewise Constant Curvature~\cite{santina2023model, hannan2003kinematics,thuruthel2019control}. While these models offer rapid predictions, they fall short in capturing complex geometries and material nonlinearities, thus sacrificing modelling accuracy. The second approach, data-driven methods, harness machine learning techniques to model and control soft robots~\cite{ Bruder2021DataDriven, Yao2024DataDriven} or deformable objects\cite{zhang2024particle, jiang2025phystwin}. Despite their potential, these approaches lack physics-based models, and require substantial data and computational resources for training robot-specific models, limiting their generalisation. The third approach includes models based on discretised continuum mechanics, such as the Finite Element Method (FEM), widely applied for their detailed and accurate representation of soft robots' actuation and geometry~\cite{largilliere2015real,grazioso2017flexible,rust2015nonlinear,duriez2013control,zhang2016kinematic,coevoet2019soft, zheng2019controllability}. FEM delivers accurate results, yet its substantial computational requirements pose challenges for integration into downstream robotics tasks.

\subsection{Differentiable FEM for Soft Robot Modelling}

Differentiable FEM promises to achieve improved balance between accuracy, efficiency, and speed~\cite{Bacher2021DesignControl}. It has already proved effective in system identification for soft robots. Dubied et al.~\cite{dubied2022sim} used differentiable FEM to estimate material parameters for three elemental soft robot configurations, employing an alternative muscle model for actuation alongside simplified geometries. Du et al.~\cite{du2021underwater} also leveraged differentiable FEM combined with a hydrodynamic model, aiming at obtaining model parameters for controlling a soft underwater robot. Their approach also incorporated an alternative elastic energy model to simulate robot actuation. However, the reliance on oversimplified models in both attempts necessitates additional manual efforts for robot-specific model adaptations. Additional efforts are sometimes also required for bridging the sim-to-real gap~\cite{gao2024sim}. Zhang et al.~\cite{zhang2022sim} addressed these limitations by generating tetrahedral meshes that accurately conform to the robot's actuation and geometry. However, this approach is limited to the relatively simple design of the robot. Meanwhile, M\'enager\cite{menager2024condensed} et al. proposed a condensed dynamics model for trajectory optimisation for soft robots, yet only demonstrated in simulation for simple trajectories. In addition, a majority of these prior approaches require dedicated motion capture setups. They also demonstrate limited exploration in leveraging the model for downstream tasks, likely due to computational intractability. These limitations considerably constrain the utility of these prior differentiable FEM approaches for soft robots beyond modelling. Our proposed framework promises to address these drawbacks and generate models for complex soft robots without oversimplified geometry or actuation mechanism. Building on top of \cite{sundaresan2022diffcloud}, we leverage point cloud observations as dense signals for system identification of a complex actuated pneumatic robot, requiring a minimal setup. Our physics-based approach also removes the need for a residual model compared to \cite{gao2024sim}, increasing generalisability and utility for downstream tasks.

\section{METHODOLOGY}

\begin{figure*} [ht]
    \vspace{2mm}
	\centering
	\includegraphics[width=0.75\textwidth]{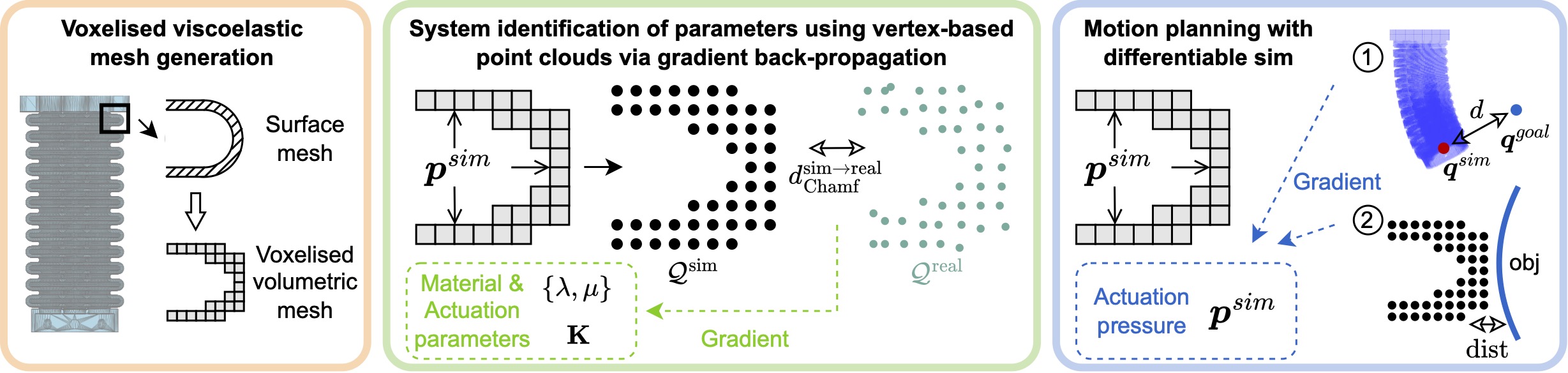}
	\caption{Overview of the methodology. Left: voxelised mesh generation from CAD. Centre: system identification of material and dynamic actuation parameters using point clouds. Right: gradient-based motion planning for path following and grasping, using distance to goal points and distance between point cloud and object as loss functions respectively.}
	\label{fig:method}
    \vspace{-5mm}
\end{figure*}

To address the challenges with soft robot modelling, this paper presents an end-to-end framework based on differentiable FEM. We begin by explaining the specifics of the FEM, the generation of the geometry model and the physics-based pneumatic actuation mechanism. We then discuss the calibration of the material and dynamic actuation parameters. Finally, we describe the trajectory optimisation techniques used to perform downstream robotics tasks. A visual overview of the methodology is in \cref{fig:method}.

\subsection{Soft Robot Modelling}
\label{sec:soft_robot_modelling}
\subsubsection{Finite Element Method (FEM)}
\label{sec:fem}
FEM is employed to simulate the dynamic behaviour of the soft robot, where we leverage tetrahedral discretisation to compute elastic forces and predict deformations of the soft body. For the differentiable FEM, we leverage commonly-used physics models and integrators with standard implementations. In modelling the viscoelastic behaviour of the tetrahedral elements, we use the Nvidia Warp off-the-shelf implementation of the Neo-Hookean constitutive model based on the strain-energy density function from Smith et al.~\cite{smith2018stable}, which yield two tunable material parameters $\lambda$ and $\mu$ that represent Lamé's first and second parameters respectively. A semi-implicit Euler integrator is then used to integrate forces on all elements to predict the motion and deformation of the finite element model, together with a penalty-based contact model. We also employ a strain-rate dissipation potential to simulate internal damping~\cite{heiden2021disect}. On top of the standard models, we designed a physics-based speed-dependent damping factor to mimic the real-world behaviour of the robot. The framework is agnostic to specific differentiable physics models. 

\subsubsection{Voxel-based Geometry Model}
To minimise manual effort, we employ an automated pipeline to generate the volumetric mesh model. Firstly, a surface mesh model is generated from the CAD model of the soft robot using standard triangulation techniques. To facilitate calibration, we first apply voxelisation to the surface mesh, which creates a voxel grid where the surface mesh is represented as cubic volumes. The empty interiors of the pneumatic chambers are preserved as unfilled voxels, while the walls are represented as filled voxels, illustrated in \cref{fig:method}. A binary-fill algorithm is employed to eliminate voids that may remain in the solid walls after voxelisation~\cite{virtanen2020scipy}. The result is a list of $N_{voxel}$ filled voxels $\{X,Y,Z\}_{N_{voxel}}$. These voxels are then converted into a tetrahedral mesh, where each voxel (cube) is subdivided into five tetrahedra using an alternating pattern based on the voxel's position. This alternation between two splitting configurations ensures isotropic behaviour in the resulting uniform tetrahedral mesh. Each tetrahedron is treated as a viscoelastic element in the FEM. As the physical robot is 3D-printed using a combination of rigid and soft materials, we initialise the rigid and soft regions with two distinct sets of material parameters, assuming uniformity within each region. For system identification, only the material parameters of the soft region are estimated.

\subsubsection{Physics-based Pneumatic Actuation} 
For a faithful physics-based model of pneumatic actuation, we identify the internal surfaces of each chamber in the soft robot and directly simulate the application of uniform pressure in each chamber. The actuation model thus requires an input pressure vector $\boldsymbol{p}^{sim}=[p_a,p_b,p_c]^T$, where the dimension corresponds to the number of chambers in the system e.g. $p_a$ refers to pressure value in chamber $a$. We define an actuation parameter set $\mathbf{K}$ to capture the mapping from real-world pressures to simulated pressures. Formally, this mapping can be expressed as $\boldsymbol{p}^{sim}=f(\boldsymbol{p}^{real};\mathbf{K})$. Here, $f(\cdot;\,\mathbf{K})$ denotes the mapping function from $\boldsymbol{p}^{real}$ to $\boldsymbol{p}^{sim}$, parameterised by $\mathbf{K}$, which ensures similar deformation behaviour between the real-world soft robot and the simulated model. It should be noted that prior analytical or differentiable FEM approaches typically struggle to directly simulate the pneumatic mechanism, especially for complex bellow-shaped soft robots.

\subsection{System Identification}
\label{sec:method:si}
After model generation, we perform identification of the material and dynamic actuation parameters. The high-fidelity geometry model enables direct comparison between the simulation and the real-world robot using point clouds.

\subsubsection{Point Cloud Generation}
\label{sec:pc_generation}
Since ground-truth mesh states of the physical robot are unavailable, we use point clouds as the observation space for system identification, inspired by DiffCloud~\cite{sundaresan2022diffcloud}. Point clouds can be easily captured using RGB-D cameras and do not require assumptions about the kinematic model of the soft robot. To compute loss in the point-cloud observation space in a differentiable manner, we directly use the vertices of the mesh as points of the point cloud. As the tetrahedral elements are derived from a voxel grid, the vertices form an evenly distributed point cloud which preserve gradient connections to the viscoelastic mesh model and its parameters, illustrated in \cref{fig:method}.

\subsubsection{Parameter Identification}
The goal of system identification is to estimate the material and dynamic actuation parameters of the simulated robot to match the behaviours of the real-world system. The material parameters include the Lamé parameters $\{\lambda,\mu\}$ for the soft regions and the actuation parameters $\mathbf{K}$, both explained in \cref{sec:fem}.

Similar to~\cite{sundaresan2022diffcloud}, we opt for the unidirectional Chamfer distance between simulated and real point clouds to compare their alignment. This mitigates some of the effects caused by sensor noise. For comparison, we crop and align the real-world and simulated point clouds using the rigidly attached top end, and project each point cloud onto a 2D plane perpendicular to the camera optical axis. The Chamfer distance is computed as ~\cref{eq:chamfer}:
\begin{equation}
    d_{\text{Chamf}}^{\text{sim} \rightarrow \text{real}} \left(\mathcal{Q}^{\text{sim}}, \mathcal{Q}^{\text{real}}\right) = 
    \sum_{\mathbf{x}^{\text{sim}} \in \mathcal{Q}^{\text{sim}}} \min_{\mathbf{x}^{\text{real}} \in \mathcal{Q}^{\text{real}}} 
    \left\| \mathbf{x}^{\text{sim}} - \mathbf{x}^{\text{real}} \right\|_2^2
\label{eq:chamfer}
\end{equation}
where $\mathbf{x}$ refers to points in point cloud $\mathcal{Q}$.
We compute loss using point clouds captured when the robot reaches steady state. This loss is then used for gradient-based optimisation of parameters through the differentiable simulation. We first optimise the material parameters $\{\lambda,\mu\}$ using observations captured when the robot is un-actuated. After convergence, we proceed to pressurise the robot at pre-defined pressure points and optimise the actuation parameters $\mathbf{K}$.

\subsection{Motion Planning}
In addition to parameter optimisation, the differentiable simulation enables gradient-based trajectory optimisation using task-space objective functions for robotics tasks such as path following and contact-based object grasping. The availability of gradients provides a guiding signal for the optimisation process, enabling efficient and locally optimal trajectory updates. Owing to the complexity, this is often computationally unstable for previous approaches to these pneumatic bellow-shaped soft robots, unless simplifying assumptions are introduced. However, by leveraging GPU-based differentiable simulation, combined with modelling strategies explained in \cref{sec:soft_robot_modelling}, the proposed framework is able to support the use of intuitive task-space metrics to optimise trajectories for downstream tasks.

\section{EXPERIMENTS}

This section details the implementation, experimental setup, and evaluation of the proposed framework for model generation and system identification for pneumatic soft robots. The experiments aim to demonstrate the effectiveness of the framework in accurately modelling the soft robot, and highlight the utility of the differentiable model for motion planning in a variety of robotics tasks.

\subsection{Implementation Details}
We demonstrate the proposed framework using a custom-designed bellow-shaped pneumatic actuator robot as a representative example. The robot CAD model is designed using SolidWorks, and the default triangulation algorithm is used to generate the surface mesh. Voxelisation is performed using the Open3D library with a voxel size of $l=0.8mm$~\cite{zhou2018open3d}. This size is chosen to be smaller than the wall thickness of the 3D-printed actuator $w=1.5mm$, ensuring that the voxelisation captures the circular geometry of the bellows. The bellow diameter of each inner chamber is $6.5mm\leq d_{chamber}\leq18.5mm$. The damping parameters are obtained by parameter sweeps for a well-damped behaviour in simulation to match the real world. The experiments are run with a quasi-static assumption. We employ Nvidia Warp as the differentiable simulation framework~\cite{warp2022}, using an RTX 6000 GPU. To avoid numerical instability due to small voxel sizes, the simulation model is uniformly scaled up by 1000 times and the parameters are identified at this scale to achieve a physically realistic behaviour (real-world observations are scaled accordingly for comparisons and loss computations). Warp's GPU-based backend enables efficient simulation of deformable objects. For point cloud loss computation, we use PyTorch and propagate gradients from PyTorch to Warp. For optimisation, we used the Adam optimiser with a learning rate of 0.01 for 100 iterations.

\subsection{Experimental Setup}
\label{sec:exp:setup}
The pneumatic actuator consists of three bellow-shaped chambers rigidly connected at both ends and linked through a central spine to minimise elongation under applied pressure. The chamber walls and central spine are 3D-printed using a soft material (Agilus30), while the ends are printed with a relatively rigid material (VeroCyan). The robot is printed using a Stratasys printer. 

\begin{figure}[!tbp]
  \centering
  \subfloat[]{\includegraphics[width=0.119\textwidth]{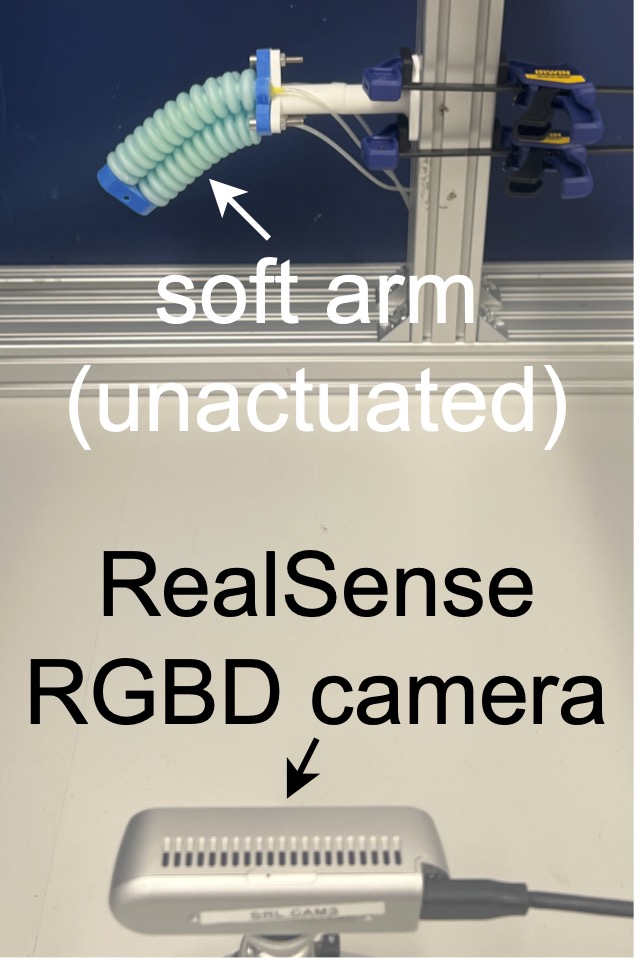}\label{fig:exp:material}}
  \hfill
  \subfloat[]{\includegraphics[width=0.352\textwidth]{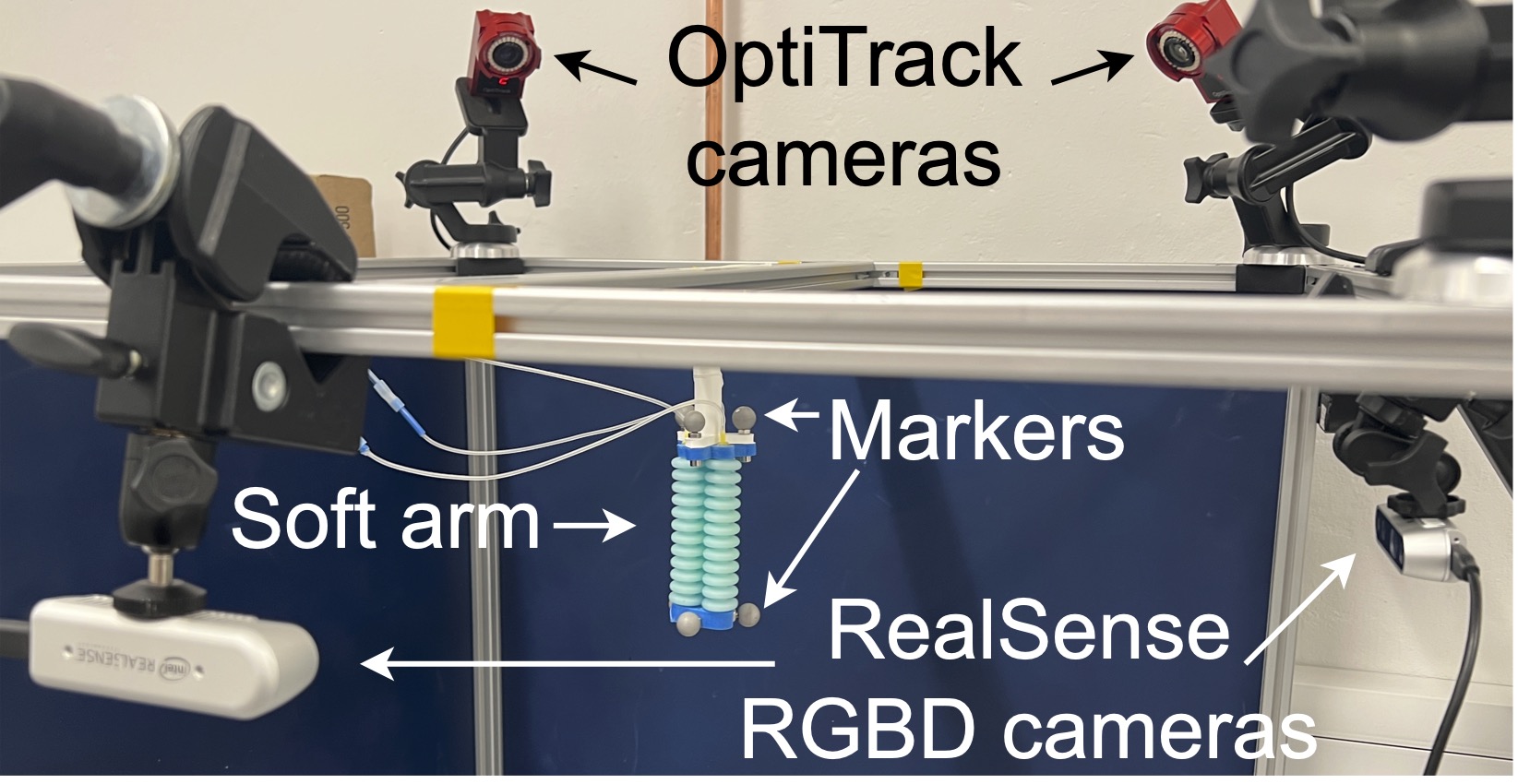}\label{fig:exp:pressure}}
  \caption{Setup for point-cloud-based system identification. a) Material parameter identification, where the robot is unactuated. b) Dynamic actuation parameter identification, where each chamber is actuated individually.}
  \vspace{-5mm}
\end{figure}

To capture point clouds for material parameter estimation, the unactuated robot is mounted on a rig, and an Intel RealSense camera is positioned to face the robot, as shown in~\cref{fig:exp:material}. For dynamic actuation parameter estimation, a separate rig is used, with two Intel RealSense cameras positioned orthogonally to capture the deformation of the actuated soft robot, as illustrated in~\cref{fig:exp:pressure}. For actuation parameter estimation, each chamber is actuated individually using the same set of pressures $p^{real}_{a,b,c}\in\{2, 4, 6, 8, 10, 12\} kPa$. Point clouds are captured once the robot reaches a steady state. Before proceeding to each pressure level, the actuator is first returned to resting position by setting all pressures to zero, mitigating potential nonlinear effects. This process yields a total of 19 data points (1 unactuated, 18 actuated) for performing system identification. For evaluation, markers are placed on the soft actuator, and their motion is tracked using OptiTrack. The average position of the three markers at the free end is used as the ground truth position $\boldsymbol{q}^{real}$.

\subsection{System Identification}
Leveraging the methods described in \cref{sec:method:si}, we perform identification of the material and dynamic actuation parameters of the soft actuator. In simulation, we run the forward simulation for 60 seconds to ensure the soft actuator reaches its steady state before comparison. 

\subsubsection{Material Parameter} 
For material parameter identification, we aim to estimate the Lamé parameters $\{\lambda,\mu\}$. The initial values of these parameters are set based on typical properties of 3D-printed materials. For optimisation, we use the point cloud of the unactuated robot attached horizontally at the top as the reference, as shown in \cref{fig:exp:material}, and configure the simulation by fixing the top end of the simulated actuator and applying the corresponding gravitational force. We then perform gradient-based updates to the material parameters to minimise the unidirectional Chamfer distance between the real-world and simulated point clouds. The material parameter identification requires only a single point cloud captured under gravity-induced deformation. Both the real-world and simulation point clouds are projected onto the same 2D plane as explained in~\cref{sec:pc_generation}.

\subsubsection{Dynamic Actuation Parameter} 
For dynamic actuation parameter identification, we aim to establish the relationship between the simulated pressure and the real-world pressure applied to each chamber of the pneumatic actuator, parameterised by $\mathbf{K}$. For each real-world pressure value, we perform gradient-based optimisation of the simulated pressure by comparing the resultant simulated and real-world point clouds at steady state. Each comparison leverages two point clouds captured from orthogonal viewing directions, as shown in \cref{fig:exp:pressure}. The point clouds are also projected onto the corresponding 2D planes for comparison as explained in~\cref{sec:pc_generation}. This set-up with two cameras is effective at addressing self-occulsions in the complex pneumatic actuator. Meanwhile the 2D projection can smooth the optimisation landscape. The Chamfer distance losses from both point cloud pairs are summed, using the loss function defined in~\cref{eq:actuation}. The parameters $\alpha$ and $\beta$, which determine the weights of the Chamfer distances, are computed based on the relative changes in the respective point clouds.
\begin{equation}
    \mathcal{L}_{act} = \alpha d_{\text{Chamf}}^{\text{sim} \rightarrow \text{real}} \left(\mathcal{Q}^{\text{sim}}_{\alpha}, \mathcal{Q}^{\text{real}}_{\alpha}\right)+\beta d_{\text{Chamf}}^{\text{sim} \rightarrow \text{real}} \left(\mathcal{Q}^{\text{sim}}_{\beta}, \mathcal{Q}^{\text{real}}_{\beta}\right)
\label{eq:actuation}
\end{equation}
Since the point clouds are captured while each chamber is pressurised individually, we update only the pressure in the corresponding chamber during optimisation, keeping the other two chambers at zero pressure.

\subsection{Evaluation}
To evaluate the model in robotics tasks, we conduct three sets of experiments. Leveraging the GPU-based differentiable simulator, we perform gradient-based trajectory optimisation using task-specific loss functions. For evaluation, the model is updated with the identified material parameters, and real-world experiments are conducted by performing sim-to-real transfer of the optimised simulation pressure values using the identified dynamic actuation parameters.

\subsubsection{Reaching} We first evaluate the accuracy of the simulation model after optimisation. Specifically, after performing system identification, we compare the predicted steady state position of the free end of the soft actuator in simulation $\boldsymbol{q}^{sim}_{T}$ with its ground-truth position in the real world $\boldsymbol{q}^{real}_{T}$, as captured by the OptiTrack system. This comparison is performed across all data points for each chamber for different pressure steps. To align the simulated and real-world movements, we normalise the origin of both positions using the initial position of the actuator's end in simulation $\boldsymbol{q}^{sim}_{0}$ and in the real world $\boldsymbol{q}^{real}_{0}$. Thus, in this way, we compare $\boldsymbol{q}^{sim}_{T}-\boldsymbol{q}^{sim}_{0}$ against $\boldsymbol{q}^{real}_{T}-\boldsymbol{q}^{real}_{0}$.

\subsubsection{Path following}
\label{sec:exp:pathfollowing}
We evaluate the differentiable model of the soft robot in a path-following task. First, we design a set of desired paths comprising way points for the free end of the soft actuator $\{ \boldsymbol{q}^{goal}_{1},...,\boldsymbol{q}^{goal}_{M}\}$. We then perform trajectory optimisation in simulation to follow these way points. The simulation length is set proportional to the number of way points $M$ to ensure that each way point can be reached in steady state. From the simulated trajectory of the free end, we sample at fixed intervals $\tau$ corresponding to each way point, and obtain $\{\boldsymbol{q}^{sim}_{\tau},...\boldsymbol{q}^{sim}_{2\tau}, \boldsymbol{q}^{sim}_{T-\tau},\boldsymbol{q}^{sim}_{T}\}$, where $T=M\tau$. For optimisation, we compute loss as the distance between each sampled position and its corresponding way point. Additionally, we use a speed-based loss to encourage the simulated robot to reach a steady state before advancing to the next way point. The loss function is defined as $
    \mathcal{L}_{path} = \sum_{m=1}^{M}d(\boldsymbol{q}^{goal}_{m},\boldsymbol{q}^{sim}_{m\tau})+\frac{1}{N} \|\mathbf{v}^{sim}_{m\tau}\|_2^2
\label{eq:path}
$%
\vspace{1mm}
where $\mathbf{v}^{sim}_{m\tau}$ represents the velocity of the point cloud $\mathcal{Q}^{\text{sim}}$ at the sampling time step $m\tau$ and $N$ is the number of points in the point cloud. Leveraging the differentiable simulation, we perform gradient-based optimisation to obtain the simulated pressure sequence $\boldsymbol{P}^{sim} = \{\boldsymbol{p}^{sim}_{1},\boldsymbol{p}^{sim}_{2},...,\boldsymbol{p}^{sim}_{M}\}$ required to achieve the desired trajectory, where $\boldsymbol{p}^{sim}=[p_a,p_b,p_c]^\top$ for the three chambers. The pressure sequence is then transferred to the real world using the actuation parameters $\mathbf{K}$, and applied on the soft actuator to achieve the corresponding real-world motion. In our experiments, $\tau$ is fixed at 20 seconds.

\subsubsection{Grasping} The high fidelity geometry and differentiability of the model provide a significant advantage compared to simplified models in enabling the computation of gradients from surface contacts. We leverage this in a grasping task by maximising contact between the surface of the simulated soft robot and a rigid rod, thereby encouraging the soft robot to grasp the rod. The loss function is defined as~\cref{eq:grasp}.
\begin{equation}
    \mathcal{L}_{grasp} = \mathcal{L}_{\text{distance}} + \mathcal{L}_{\text{contact}}
\label{eq:grasp}
\vspace{-2.5mm}
\end{equation}
where 
\vspace{-2.5mm}
\begin{align}
    \mathcal{L}_{\text{distance}} &= \frac{1}{N} \sum_{i=1}^{N} \text{dist}_i^2
    \\
    \mathcal{L}_{\text{contact}} &= -\frac{1}{N} \sum_{i=1}^{N} I(\epsilon - \text{dist}_i)
\label{eq:dist_cont}
\end{align}
$\text{dist}_i$ computes distance of individual points in the point cloud $\mathcal{Q}^{\text{sim}}$ to the rod, while $I(\epsilon - \text{dist}_i)$, implemented using a differentiable selection operation, returns a value of 1 for a distance within a threshold $\epsilon$, indicating contact. This ensures contact detection is differentiable. Using the model, pressure $\boldsymbol{p}^{sim}$ can be optimised through gradients directly computed from this straightforward contact-based loss function.

\section{RESULTS AND ANALYSIS}
In this section, we present and analyse the results of model generation, system identification, and model evaluation in downstream robotics tasks.

\subsection{Soft Robot Modelling}
\label{sec:result:model}
\begin{figure}[htbp]
  \centering
  {\includegraphics[width=0.2375\textwidth]{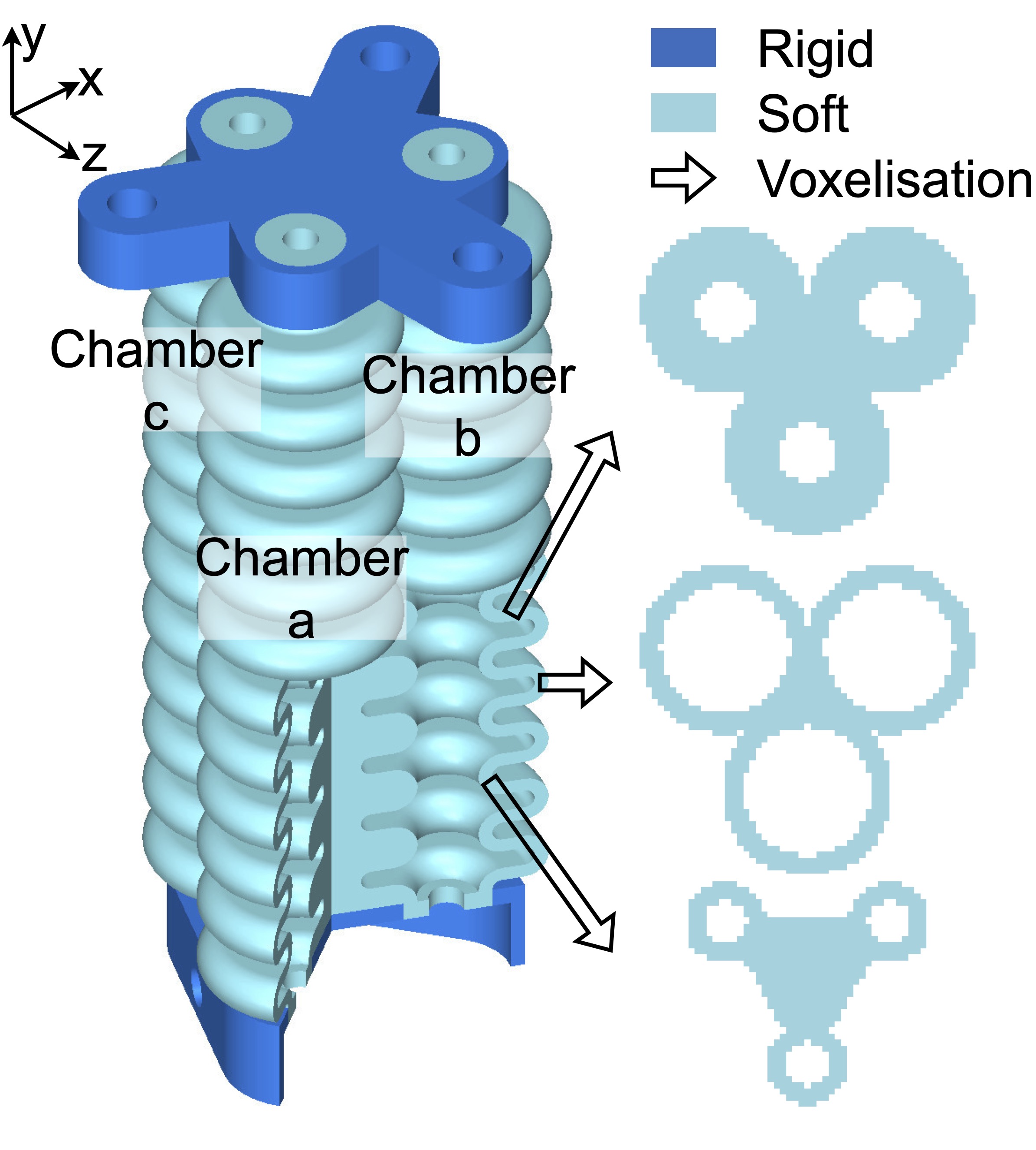}}
  \caption{Illustration of the soft actuator. The cross-sections indicate that the voxelisation pipeline can maintain the shapes of the internal chambers.}
  \label{fig:cross}
  \vspace{-4mm}
\end{figure}

After voxelisation of the surface mesh of the robot, we obtain a total of $\sim161,000$ filled voxels, which results in $\sim805,000$ tetrahedra elements. These elements share $\sim211,000$ vertices, which collectively form the simulation point cloud $\mathcal{Q}^{\text{sim}}$. As shown in \cref{fig:cross}, the voxelisation process preserves the circular inner chambers of the bellow-shaped actuator. By applying forces to the faces of these inner chambers, we can simulate pneumatic actuation in an accurate physics-based manner. The prior approaches are unable to achieve this with their oversimplified models. In our work, this is enabled by the high fidelity geometry model powered by GPU-based simulation. The subsequent results demonstrate the utility of the model for system identification using point clouds and for downstream robotics tasks. Furthermore, on GPU, we are able to achieve an average of more than 3000 FPS on forward simulation and more than 160 FPS on backward computation. When using only the CPU, the forward simulation runs at less than 15 FPS and the backward computation at less than 2 FPS. This demonstrates the advantage in leveraging GPU acceleration for the physics and gradient computation.
\vspace{-2mm}

\subsection{System Identification}
The proposed framework enables the estimation of material and dynamic actuation parameters of the differentiable model using easily captured point clouds. By comparing the pair of point clouds captured when the robot deforms under gravity, we obtain $\{\lambda=1.4 \times 10^{8} \text{ Pa},\mu=7.27 \times 10^{7} \text{ Pa}\}$, which fall within the expected range for the model of the soft material used in the 3D-printed soft robot.

\begin{figure} [h!]
\vspace{1mm}
	\centering
	\includegraphics[width=0.525\columnwidth]{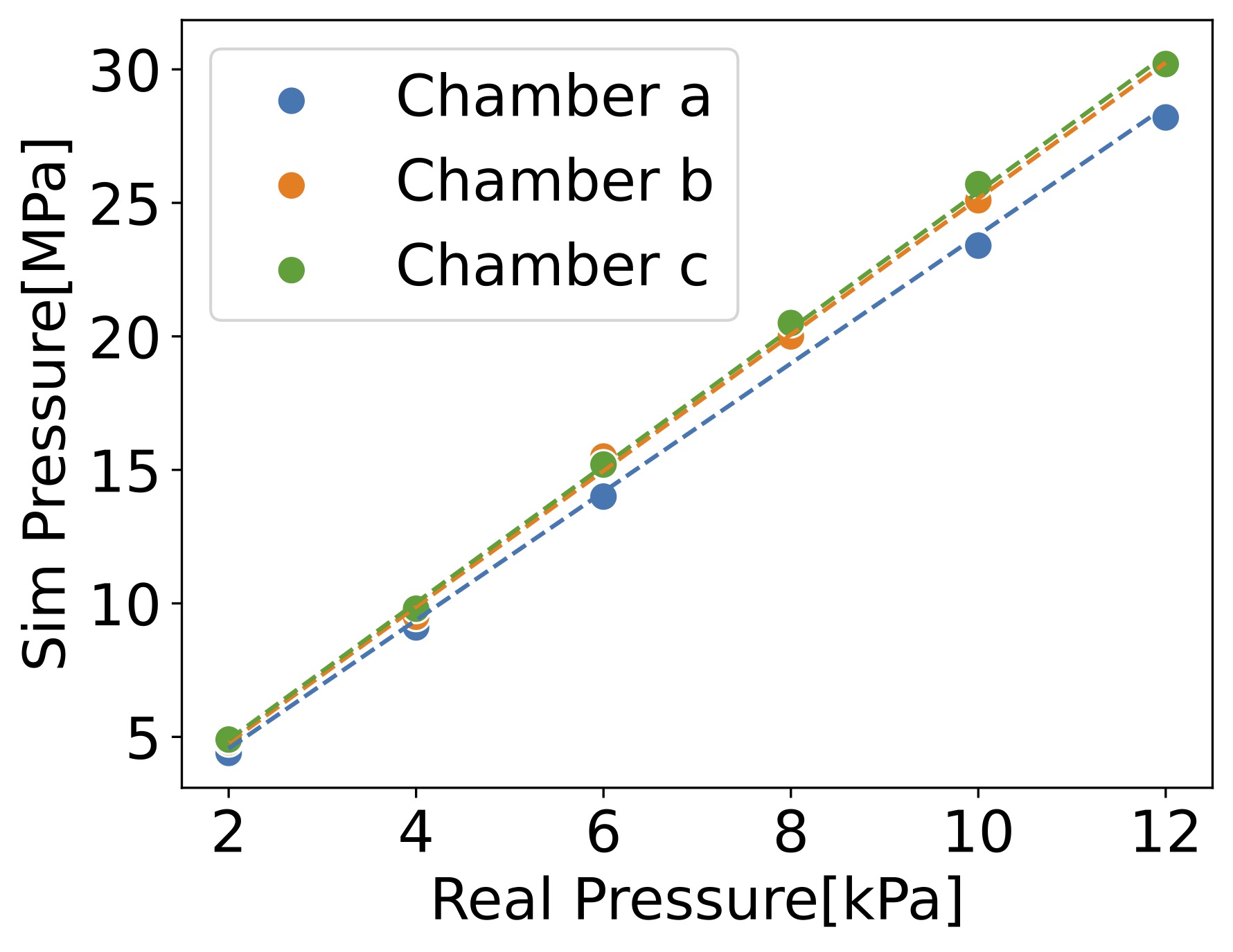}
	\caption{Simulation pressure vs real pressure. The linear relationship between the simulated and real-world pressure for each chamber demonstrates that the model effectively captures the nonlinear dynamics of the soft robot.}
	\label{fig:calib}
\vspace{-4.5mm}
\end{figure}

To estimate the dynamic actuation parameters $\mathbf{K}$, we leverage point clouds captured when the robot is actuated at different pressure values. For each pressure $p^{real}$ applied to each chamber, point-cloud-based optimisation converges to a simulated pressure $p^{sim}$ that results a simulated deformation that best mimics the robot's real-world behaviour. As described in \cref{sec:exp:setup}, we obtain six $\{p^{real},p^{sim}\}$ pairs for each chamber $\{a,b,c\}$. \cref{fig:calib} exhibits a linear relationship between the simulated and real-world pressures for all chambers. This shows that a linear mapping from the real pressure to the simulated pressure is sufficient to achieve a similar deformation behaviour between the simulated model and the physical robot, demonstrating that the differentiable model is able to capture the nonlinear dynamics of the soft actuator. Thus, we model the pressure relationship as:
\vspace{-0.5mm}
\begin{equation}
    \boldsymbol{p}^{sim} = \boldsymbol{\kappa}\boldsymbol{p}^{real}+\boldsymbol{\rho}
\label{eq:pressure}
\end{equation}
\vspace{-0.5mm}
where $\mathbf{K}=\{\boldsymbol{\kappa},\boldsymbol{\rho}\}$, $\boldsymbol{\kappa}$ is a diagonal matrix with elements $\mathrm{diag}(\kappa_a, \kappa_b, \kappa_c)$, and $\boldsymbol{\rho}=[\rho_a,\rho_b,\rho_c]^T$. This validates the effectiveness of the point-cloud-based approach for system identification of the bellow-shaped pneumatic soft robot.

\subsection{Reaching}
\begin{figure} [h!]
	\centering
	\includegraphics[width=0.525\columnwidth]{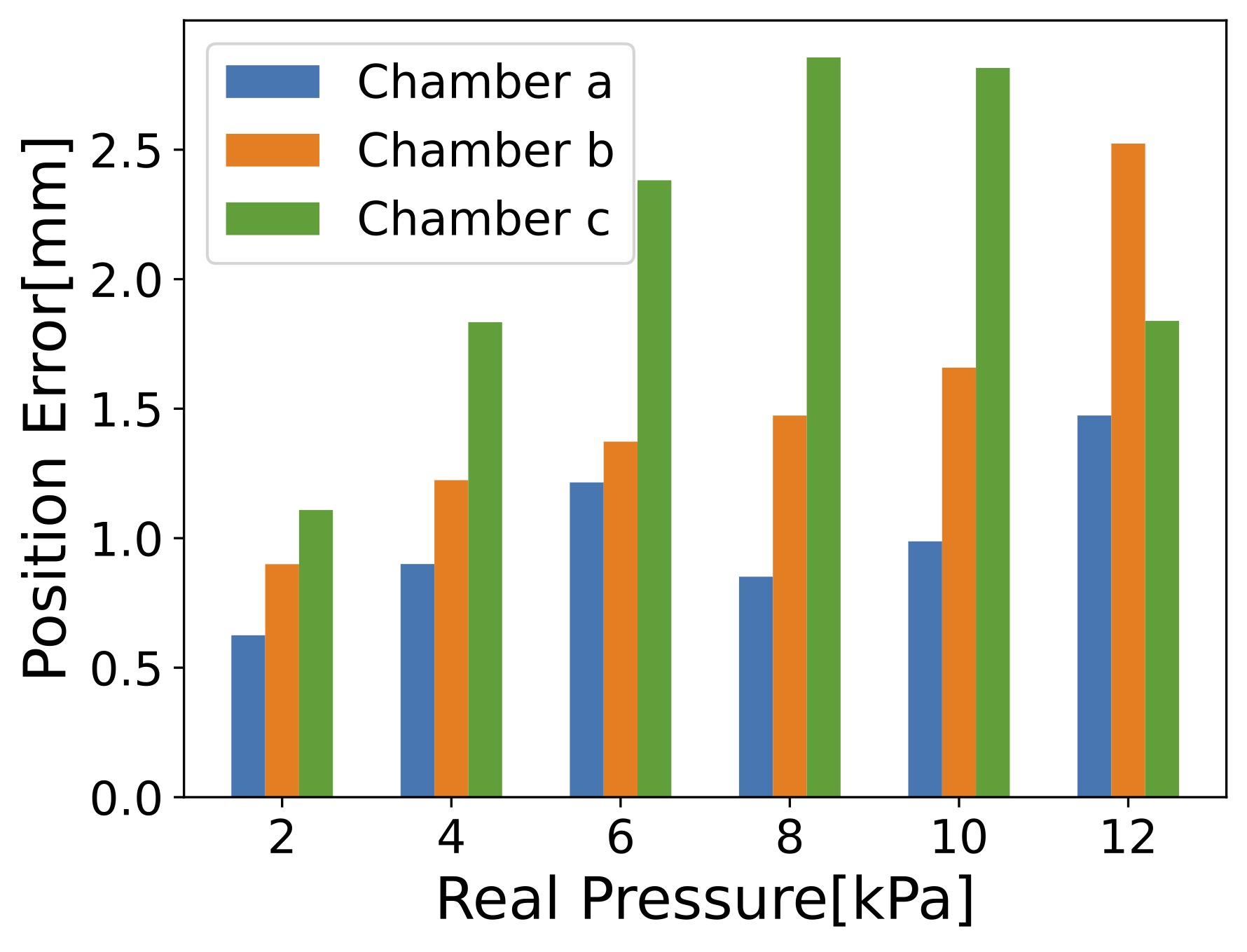}
	\caption{Position error for each chamber at different real-world pressures. The error is less than 3\% of the robot's length for all chambers, demonstrating high modelling accuracy.}
	\label{fig:ee}
    \vspace{-4.5mm}
\end{figure}

After obtaining the dynamic actuation parameters, we compare the simulated and real-world movements of the free end of the soft actuator at corresponding pressure pairs $\{p^{real},p^{sim}\}$. As shown in~\cref{fig:ee}, the model achieves a small position error across different pressures, with the maximum error being less than $3mm$, representing less than 3\% of the actuator's length. This accuracy is comparable to state-of-the-art soft robot modelling approaches, validating the model's ability to simulate real-world behaviour using parameters identified from point clouds. 

Notably, this performance is achieved through direct physics-based modelling of pneumatic actuation within the complex bellow-shaped chambers, without additional actuation mechanisms or manually introduced biases. Additionally, the geometry modelling enables the generation of simulated point clouds that closely match real-world observations and facilitate Chamfer distance computation. As expected, errors generally increase at higher pressures, where the actuator undergoes larger, more nonlinear deformations. The error is slightly higher for \texttt{Chamber c}, likely due to its placement behind the other two chambers, which obstructs one camera's view and affects the optimisation results.
\vspace{-2mm}

\subsection{Path Following}
\begin{figure*}[tp]
  \centering
  \subfloat[Square]{\includegraphics[width=0.23\textwidth]{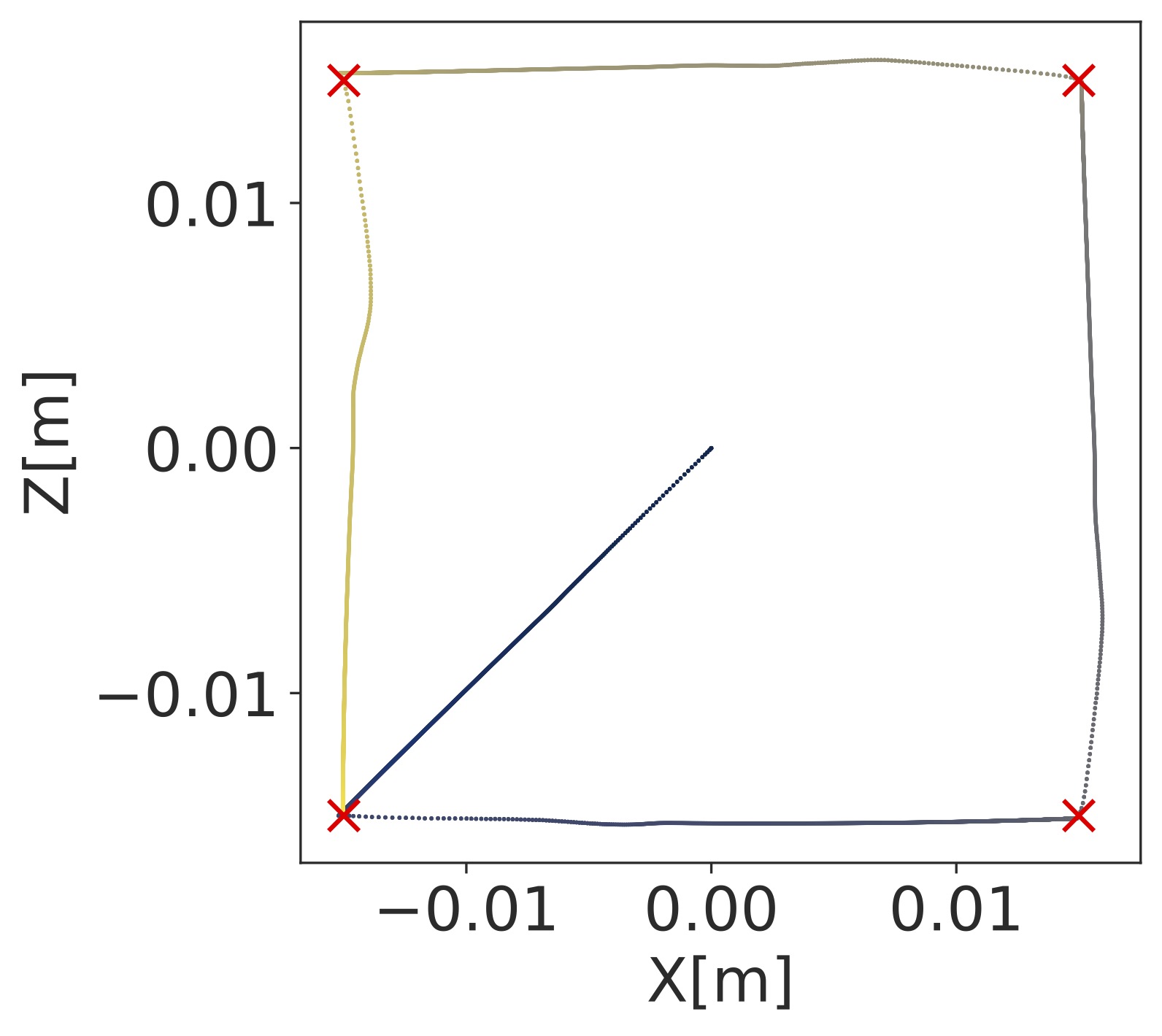}\label{fig:sim:square}}
  \hfill
  \subfloat[Triangle]{\includegraphics[width=0.26\textwidth]{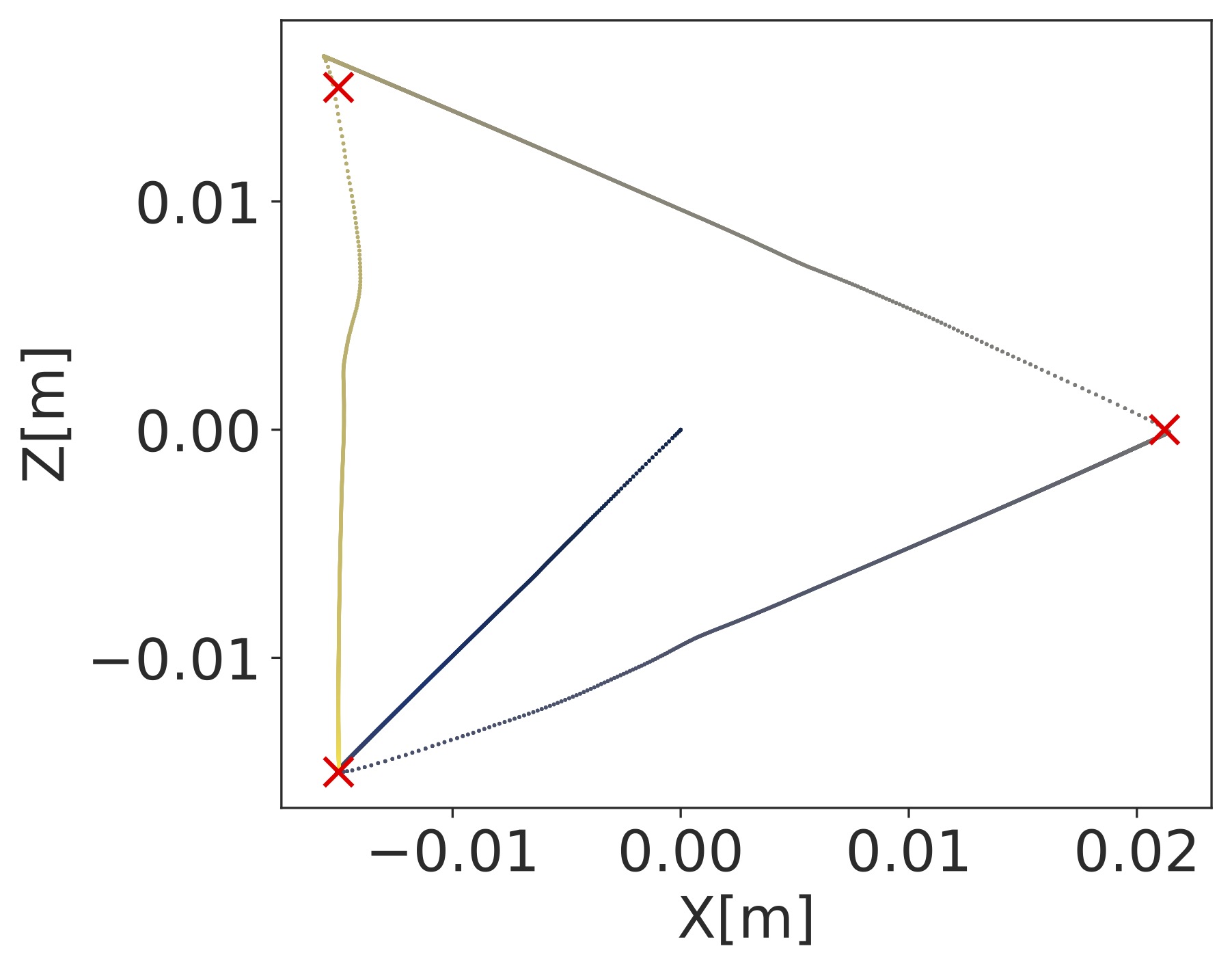}\label{fig:sim:tri}}
  \hfill
  \subfloat[Eight]{\includegraphics[width=0.28\textwidth]{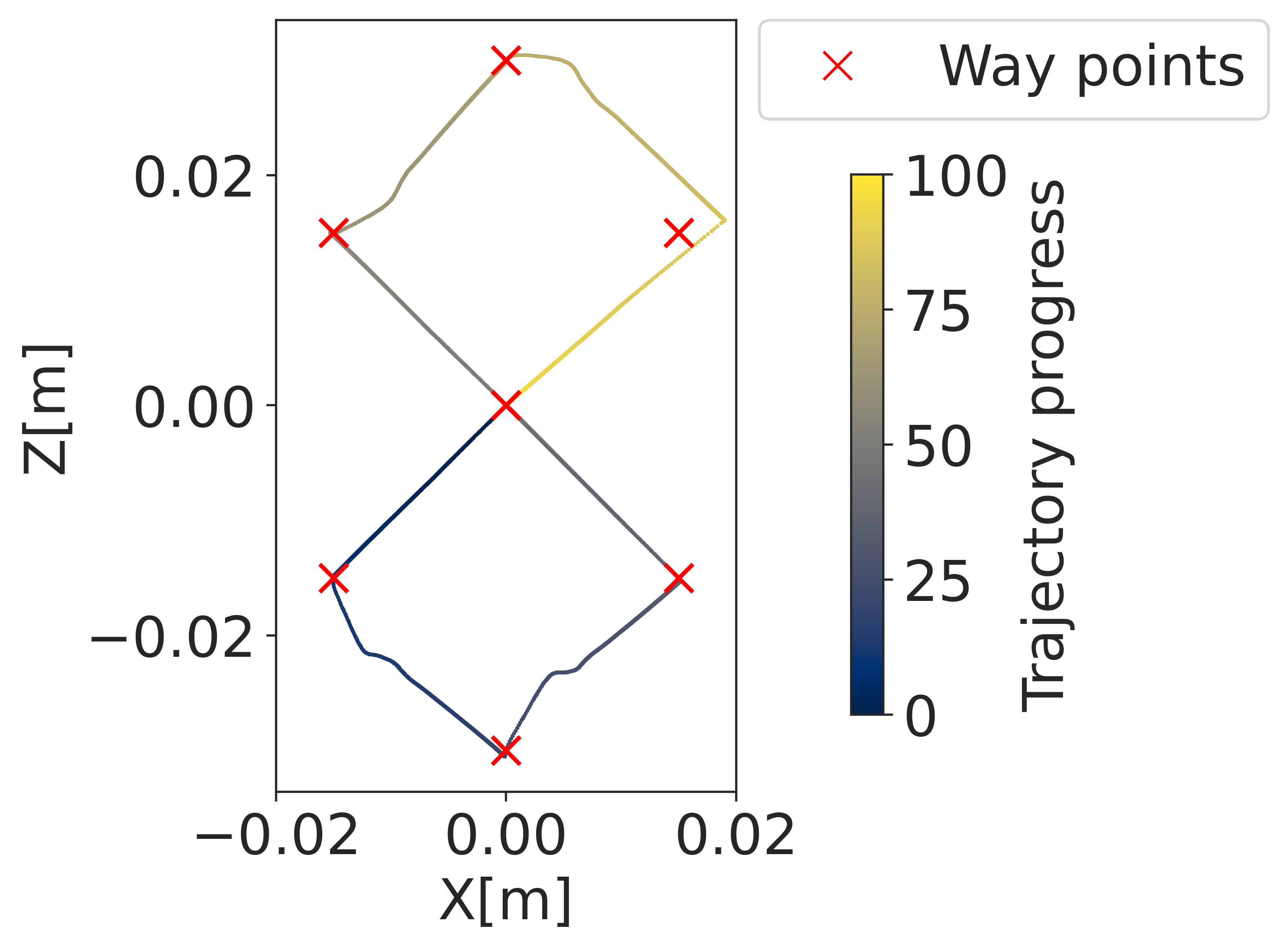}\label{fig:sim:eight}}
\vspace{-1mm}
  \caption{Path following results in simulation. Each path is executed 50 times with results overlaid. This demonstrates that the simulated robot successfully follows the desired paths, highlighting the model's efficacy for trajectory optimisation.}
  \label{fig:sim}
  \vspace{-5mm}
\end{figure*}

\begin{figure*}[tp]
  \centering
  \subfloat[Square]{\includegraphics[width=0.225\textwidth]{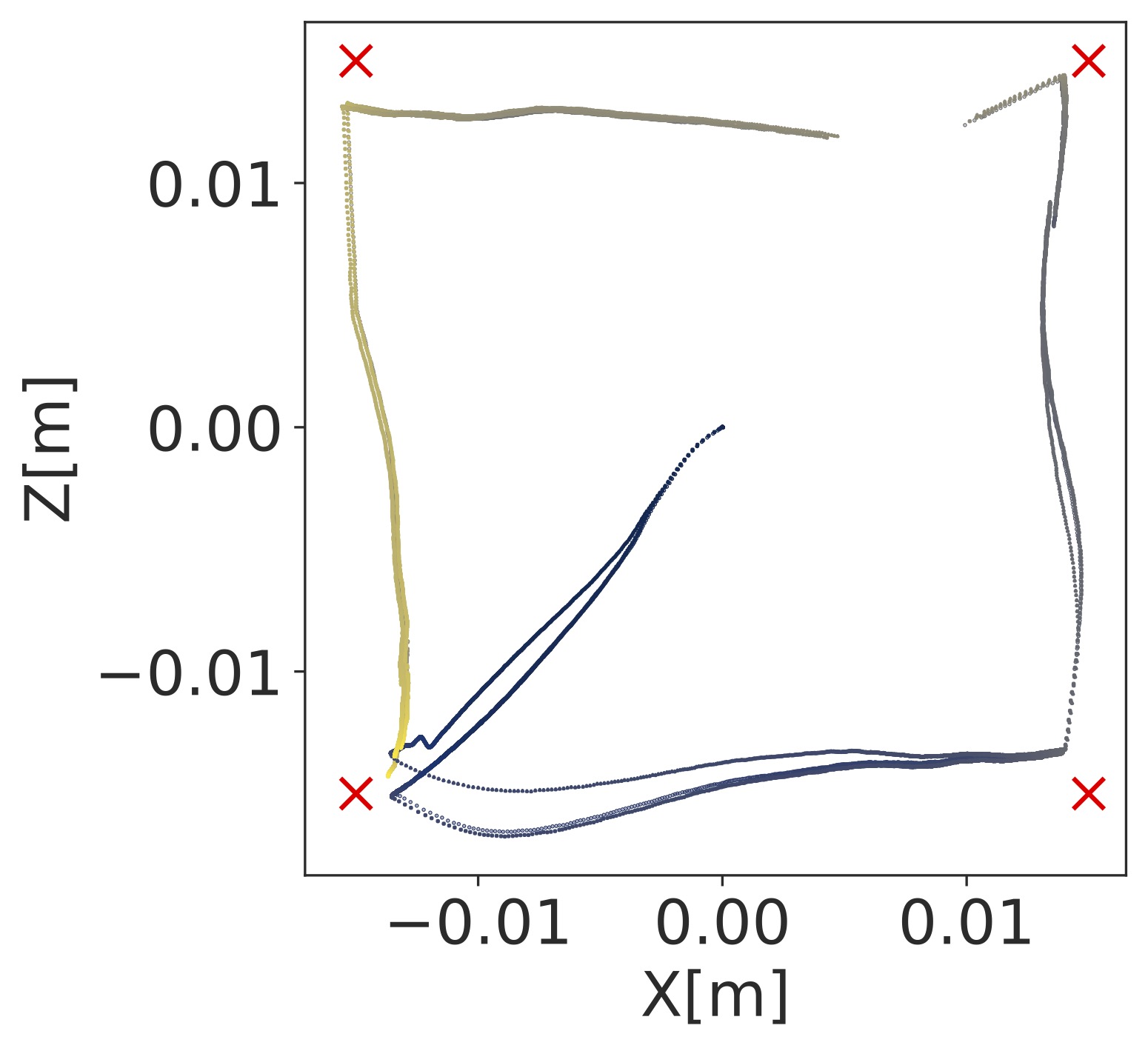}\label{fig:real:square}}
  \hfill
  \subfloat[Triangle]{\includegraphics[width=0.275\textwidth]{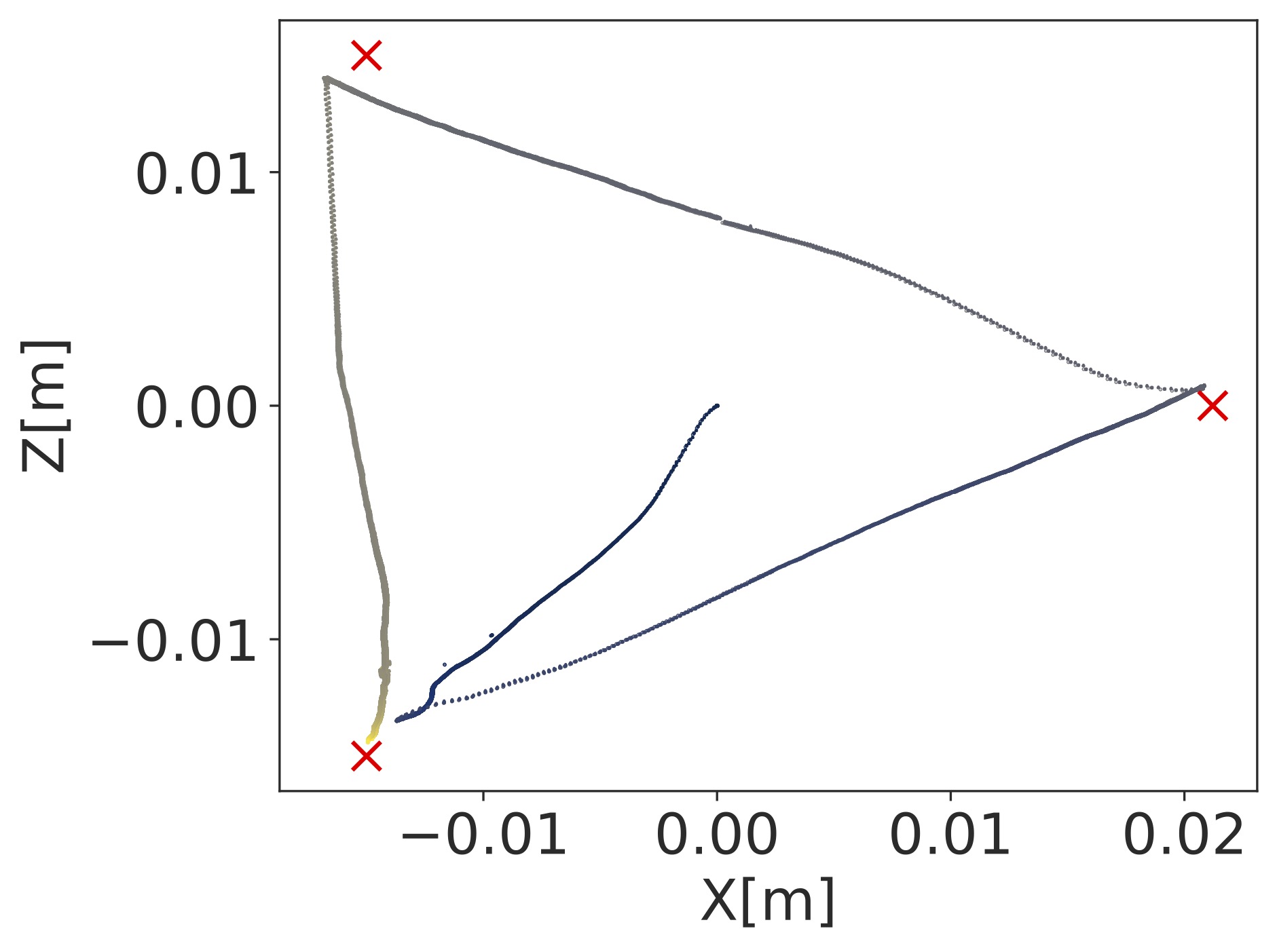}\label{fig:real:tri}}
  \hfill
  \subfloat[Eight]{\includegraphics[width=0.28\textwidth]{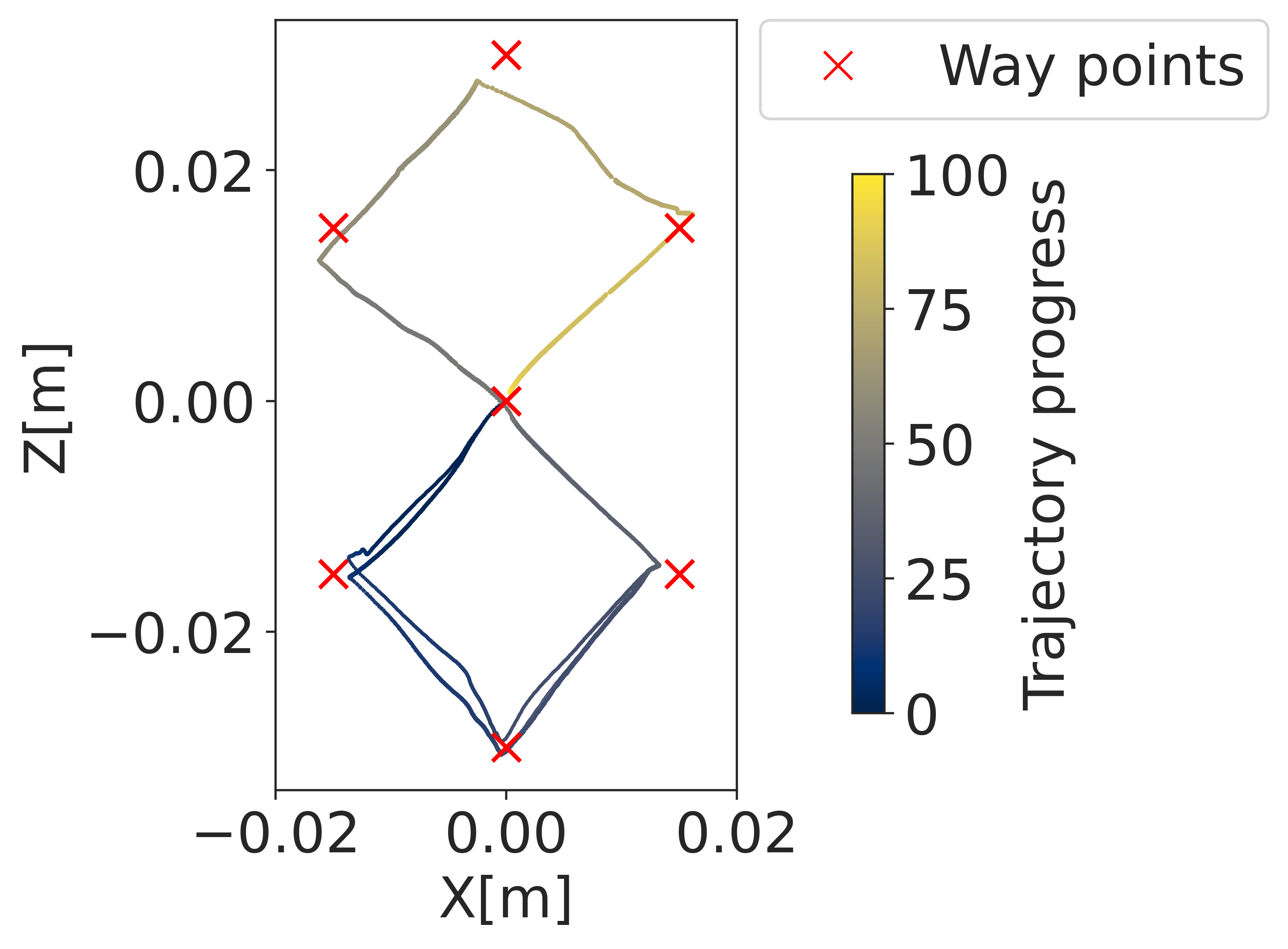}\label{fig:real:eigth}}
\vspace{-1mm}
  \caption{Path following results in the real world. Each path is executed 3 times with results overlaid. This demonstrates successful sim-to-real transfer using the identified parameters. The gap in the square trajectory is likely due to occlusions.}
  \label{fig:real}
  \vspace{-4mm}
\end{figure*}

We evaluate the differentiable model of the soft robot in a path-following task. Using the pipeline described in \cref{sec:exp:pathfollowing}, we obtain a pressure sequence that actuates the simulation model to follow a sequence of way points along three reference paths. The simulation results, shown in \cref{fig:sim}, demonstrate that the model accurately follows target paths via specified way points. All target way points are reached with an average error of 0.1mm. The average deviation from target paths (the mean distance between the simulated trajectory and the continuous straight path connecting target way points) is less than 0.5mm, indicating the convergence of optimisation. It is worth noting that only a small number of way points are used (5 for the square, 4 for the triangle, and 8 for the figure-of-eight), and additional way points can be included to achieve even more precise path tracking. These results highlight the capability of the physics-based pneumatic actuation model to achieve desired motions for arbitrary paths and underscore the utility of the model in gradient-based trajectory optimisation for soft robots.

After performing optimisation in simulation, we transfer the pressure sequences to the real world using the identified dynamic actuation parameters $\mathbf{K}$ and apply them to the physical robot. The results, shown in \cref{fig:real}, demonstrate successful sim-to-real transfer, enabling the robot to follow arbitrary target paths. The target way points are reached with an average error of 1.2mm. The average deviation from target path is less than 0.9mm, with maximum below 3.2mm, commensurate with the reaching results. Notably, the dynamic actuation parameters are identified for each chamber individually, yet the path-following task requires pressurising different combinations of chambers simultaneously, introducing nonlinear interactions as the actuation of one chamber affects the others. The results indicate that the simulation model effectively handles these nonlinear effects and can be applied to real-world soft robotics tasks. To the best of the authors' knowledge, this framework is the first to extend gradient-based trajectory optimisation to such complex bellow-shaped pneumatic soft robots, addressing the challenges of modelling their complex geometry, dynamics, and actuation mechanisms. 

\subsection{Grasping}
\label{sec:results:grasp}
\begin{figure}[hbp]
  \vspace{-1mm}
  \centering
  {\includegraphics[width=0.45\textwidth]{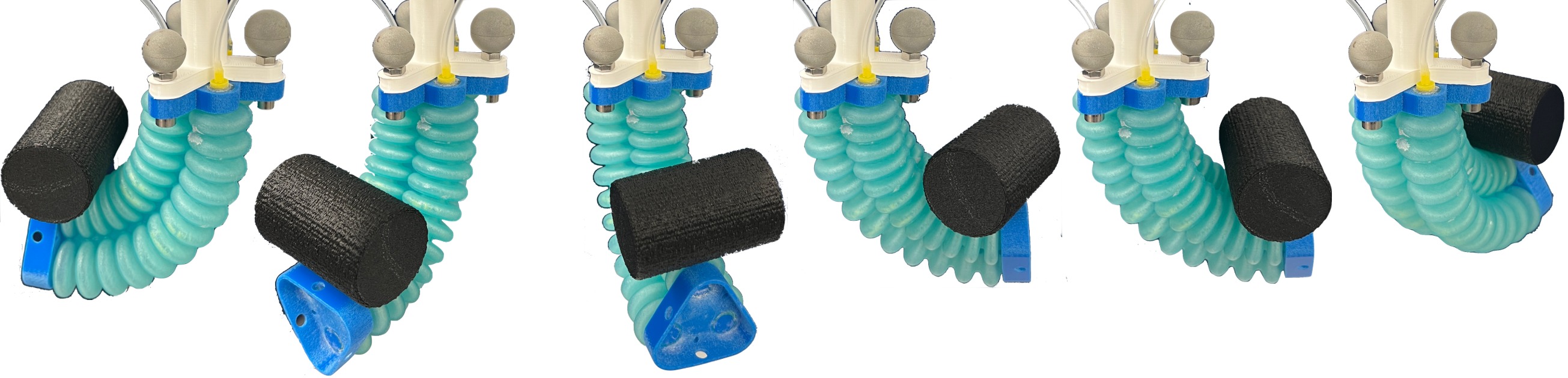}}
  \caption{Illustration of grasping. The simple loss function for maximising contact enables successful grasping at different poses. This shows an ability for trajectory optimisation for contact-based tasks and generalisation to object poses.}
  \label{fig:grasp}
  \vspace{-2mm}
\end{figure}

We further demonstrate the utility of the differentiable model in a grasping task that involves contacts with an object at different poses. Using the loss function in ~\cref{eq:grasp}, we optimise for pressure $\boldsymbol{p}^{sim}$ applied on the three chambers of the soft actuator to maximise contact with a rod object placed at various poses around the robot model in simulation. After convergence, the pressure is transferred to the real world using the actuation parameters $\mathbf{K}$. As shown in \cref{fig:grasp}, through the proposed framework, the simple loss function for maximising contact enables successful grasping and generalises to different object poses. This capability is enabled by the accurate geometry modelling of the soft robot, which allows for precise detection of contacts at the surfaces of the bellow-shaped structure. Modelling such environmental interactions has traditionally been intractable for prior methods due to the complexity of coupling with the environment and the high dimensionality of the soft robot. Yet the proposed framework demonstrates the ability to intuitively account for contacts and perform trajectory optimisation for such robots in these contact-based tasks.

The evaluation results demonstrate the effectiveness of the end-to-end framework for modelling pneumatic soft robots. Using a CAD model as input, the framework generates a functional differentiable finite element model that incorporates physics-based pneumatic actuation and can be easily calibrated using point clouds. Once calibrated, the model demonstrates efficacy in a variety of real-world robotics tasks through gradient-based trajectory optimisation. This is done without robot or task specific modifications. The framework is thus task- and robot-agnostic, suggesting its potential for generalisation to other tasks and pneumatic robots.

\section{CONCLUSIONS}
In this paper, we propose an end-to-end framework for modelling pneumatic soft robots using differentiable FEM. We leverage an automated voxelisation pipeline for model creation, and use point clouds for system identification. We demonstrate the framework's effectiveness using a complex bellow-shaped pneumatic soft actuator and validate the model through a series of downstream robotics tasks. The approach enables accurate, efficient, physics-based modelling of the soft robot and requires minimal human efforts. The results show that the framework accurately captures the robot's geometry, dynamics, and actuation mechanisms, and that the generated model can be applied to various robotics tasks. Future work should extend the framework to robots with different actuation mechanisms and material compositions, address the computational requirement, and explore different differentiable physics models to reduce the potential for numerical instability.

\section*{ACKNOWLEDGMENT}

The authors would like to acknowledge the use of the University of Oxford Advanced Research Computing (ARC) facility http://dx.doi.org/10.5281/zenodo.22558 and the SCAN facility. Ingmar Posner holds concurrent appointments as a Professor of Applied AI at the University of Oxford and as an Amazon Scholar. This paper describes work performed at the University of Oxford and is not associated with Amazon.

\bibliographystyle{IEEEtran}
\bibliography{library}
\end{document}